\documentclass[sigconf]{acmart}

\AtBeginDocument{%
  \providecommand\BibTeX{{%
    \normalfont B\kern-0.5em{\scshape i\kern-0.25em b}\kern-0.8em\TeX}}}

\setcopyright{acmcopyright}
\copyrightyear{2024}
\acmYear{2024}
\acmDOI{XXXXXXX.XXXXXXX}

\acmPrice{15.00}
\acmISBN{978-1-4503-XXXX-X/18/06}

\usepackage{subfigure}
\usepackage{amsthm}
\usepackage{booktabs}
\usepackage{multirow}
\usepackage{graphicx}
\graphicspath{{./figs/}}
\usepackage[linesnumbered,ruled,vlined]{algorithm2e}
\usepackage{color}
\usepackage{enumitem}

\begin{document}

\title{Enhance Graph Alignment for Large Language Models}

\author{Haitong Luo}
\affiliation{%
  \institution{Institute of Computing Technology, Chinese Academy of Sciences}
  \city{Beijing}
  \country{China}
}
\author{Xuying Meng}
\affiliation{%
  \institution{Institute of Computing Technology, Chinese Academy of Sciences}
  \city{Beijing}
  \country{China}
}
\author{Suhang Wang}
\affiliation{%
  \institution{Pennsylvania State University}
  \city{State College}
  \country{USA}
}
\author{Tianxiang Zhao}
\affiliation{%
  \institution{Pennsylvania State University}
  \city{State College}
  \country{USA}
}
\author{Fali Wang}
\affiliation{%
  \institution{Pennsylvania State University}
  \city{State College}
  \country{USA}
}
\author{Hanyun Cao}
\affiliation{%
  \institution{Institute of Computing Technology, Chinese Academy of Sciences}
  \city{Beijing}
  \country{China}
}
\author{Yujun Zhang}
\affiliation{%
  \institution{Institute of Computing Technology, Chinese Academy of Sciences}
  \city{Beijing}
  \country{China}
}
\begin{abstract}

Graph-structured data is prevalent in the real world. Recently, due to the powerful emergent capabilities, Large Language Models (LLMs) have shown promising performance in modeling graphs. The key to effectively applying LLMs on graphs is converting graph data into a format LLMs can comprehend. Graph-to-token approaches are popular in enabling LLMs to process graph information. They transform graphs into sequences of tokens and align them with text tokens through instruction tuning, where self-supervised instruction tuning helps LLMs acquire general knowledge about graphs, and supervised fine-tuning specializes LLMs for the downstream tasks on graphs. Despite their initial success, we find that existing methods have a misalignment between self-supervised tasks and supervised downstream tasks, resulting in negative transfer from self-supervised fine-tuning to downstream tasks. To address these issues, we propose Graph Alignment Large Language Models (GALLM) to benefit from aligned task templates. In the self-supervised tuning stage, we introduce a novel \textit{text matching} task using templates aligned with downstream tasks. In the task-specific tuning stage, we propose two category prompt methods that learn supervision information from additional explanation with further aligned templates. Experimental evaluations on four datasets demonstrate substantial improvements in supervised learning, multi-dataset generalizability, and particularly in zero-shot capability, highlighting the model's potential as a graph foundation model.

\end{abstract}






\maketitle

\section{Introduction}

The Web has become a universal data repository, linking diverse entities to form intricate graphs. This graph-structured data is crucial in our daily lives, driving applications like Web mining \cite{agarwal2022graphnli,xu2022evidence}, content recommendation \cite{yang2021consisrec,luo2024spectral}, and anomaly detection \cite{liu2021pick,tang2022rethinking}. While graphs effectively represent complex interactions, they pose significant challenges in analysis and modeling due to their inherent irregularity and diverse connectivity patterns.
Recently, Large Language Models (LLMs), such as GPTs \cite{achiam2023gpt}, LLaMA \cite{touvron2023llama} and Claude \cite{perez2022discovering}, have emerged as a dominant force in AI research due to their remarkable generative ability. 
Trained on vast amounts of text data, these models have demonstrated exceptional performance not only in natural language processing tasks \cite{devlin2019pre, raffel2020exploring}, but also in fields beyond texts such as computer vision \cite{yin2023survey}, recommendation systems \cite{bao2023tallrec}, and speech recognition \cite{fathullah2024prompting}.
The emergent capabilities of LLMs also inspire researchers to apply LLMs to text-attributed graphs, aiming to develop a graph foundation model that functions effectively across multiple scenarios \cite{liu2023towards}. The key to employing LLMs for graph data lies in aligning the graph data with natural language/text embedding space so that LLMs can comprehend the graph \cite{liu2023towards}. 

Several efforts have been taken on the alignment of graph data, which can be generally categorized into two categories: graph-to-text \cite{liu2023evaluating, wang2024can, zhao2023graphtext, guo2023gpt4graph, chen2024exploring, fatemi2023talk} and graph-to-token \cite{ye2023natural,tang2024graphgpt,chen2024llaga, kong2024gofa} approaches. \textit{Graph-to-text} methods use natural language to describe graphs. However, such simple strategies often suffer from redundancy and imprecise characterization due to the manual setup of description/prompt templates. 
In contrast, \textit{graph-to-token} methods convert graph information into sequences of tokens and then obtain their embeddings. They treat these graph tokens as a new modality, and project these tokens to align the feature space with text embeddings, facilitating the understanding of LLM on graphs. 

Though graph-to-token methods alleviate the imprecise characterization issue, existing works are suboptimal for alignment. To align new modality tokens, a two-stage instruction tuning paradigm serves as a prevalent strategy in existing works \cite{ye2023natural, tang2024graphgpt}: (i) In Stage 1 (i.e., the self-supervised tuning stage), LLMs are fine-tuned using self-supervised tasks to align new modality tokens with text tokens, allowing them to acquire more domain-related knowledge without being constrained by the amount of supervised data; and (ii) In Stage 2 (i.e., the task-specific tuning stage), the LLM is fine-tuned on specific downstream tasks, thereby infused with task-related knowledge. For example, InstructGLM \cite{ye2023natural} employs a self-supervised link prediction task to enhance the downstream node classification tasks. GraphGPT \cite{tang2024graphgpt} proposes \textit{graph matching} tasks in Stage 1 to boost the performance of downstream node classification and link prediction tasks. LLaGA \cite{chen2024llaga} adopts \textit{node description} as the self-supervised task. However, the above methods all fall short in the two-stage instruction tuning paradigm, as {they overly rely on supervised task-specific tuning and struggle to benefit from self-supervised tuning. 

Self-supervised tasks proposed by existing methods \cite{ye2023natural, tang2024graphgpt, chen2024llaga} differ significantly from downstream tasks, making it challenging for models to benefit from self-supervised tuning, and even leading to negative transfer \cite{wang2021afec}. As shown in Figure \ref{fig:prelinminary_study}, we conduct a 
preliminary study to compare performance based on different self-supervised tasks, with Vicuna-7B-v1.5-16K as the base LLM model on Cora \cite{yang2016revisiting} and Pubmed \cite{he2023explanations} datasets. 
Except for the base model Vicuna without self-supervised tuning, we utilize two abovementioned representative self-supervised tasks, i.e., graph matching (\textit{w/} gm) \cite{tang2024graphgpt} and node description (\textit{w/} nd) \cite{chen2024llaga}.
All models are instruction-tuned for the node classification task, differing in their self-supervised tuning stages.
From Figure \ref{fig:prelinminary_study}, we can observe that the performance of the models decreases after self-supervised tuning with either node description or graph matching. This negative transfer phenomenon occurs due to the inconsistency between the templates and training objectives of self-supervised tasks and those of the downstream tasks. Specifically, node description is a generative task that requires generating descriptive text for nodes, while graph matching involves reordering nodes based on given texts. In contrast, the downstream node classification task is a discriminative task that requires the LLM to select the correct category from a set of candidate categories.

\begin{figure}[t]
    \centering{
    \includegraphics[scale=0.082]{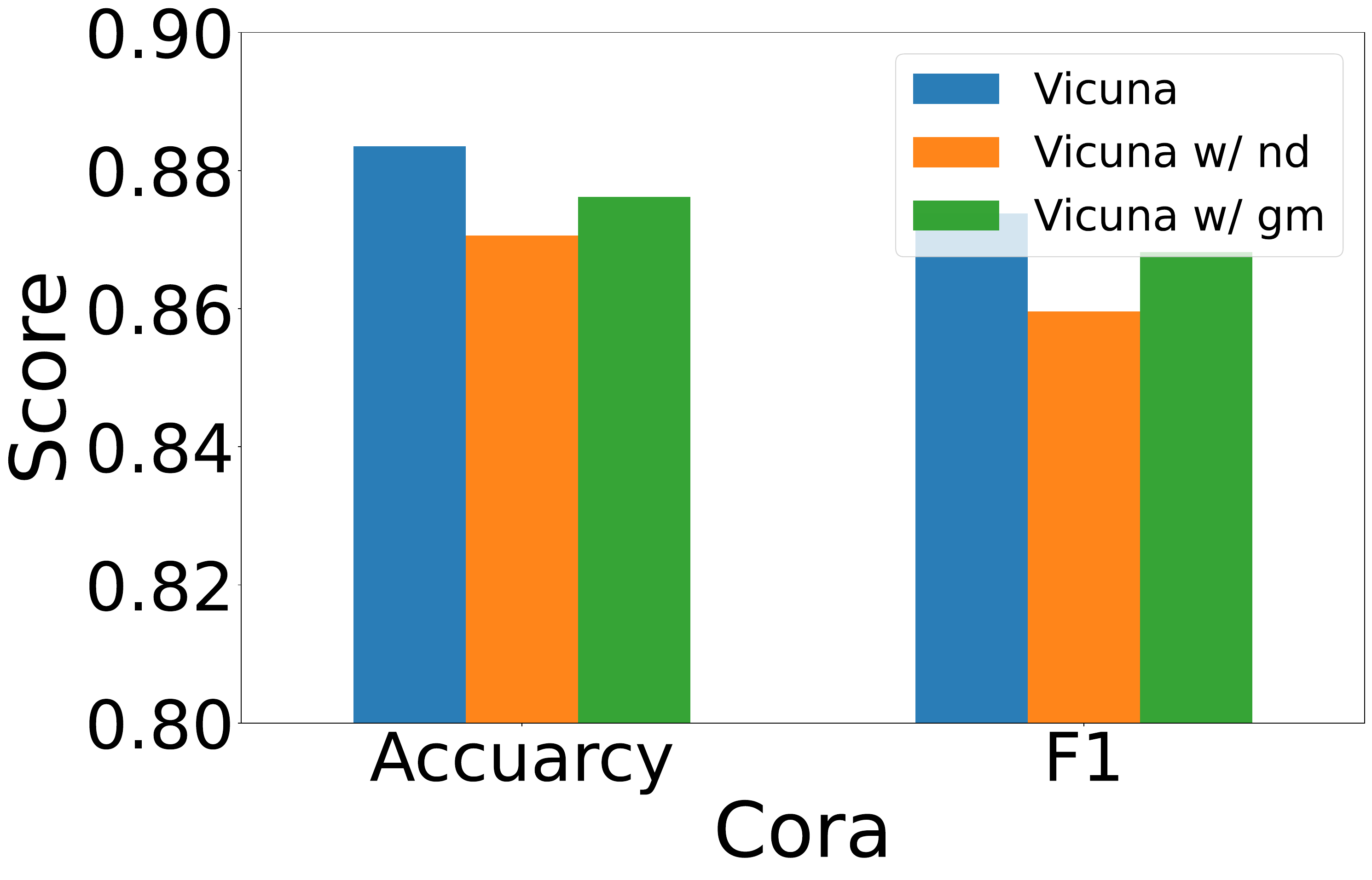}
    \includegraphics[scale=0.082]{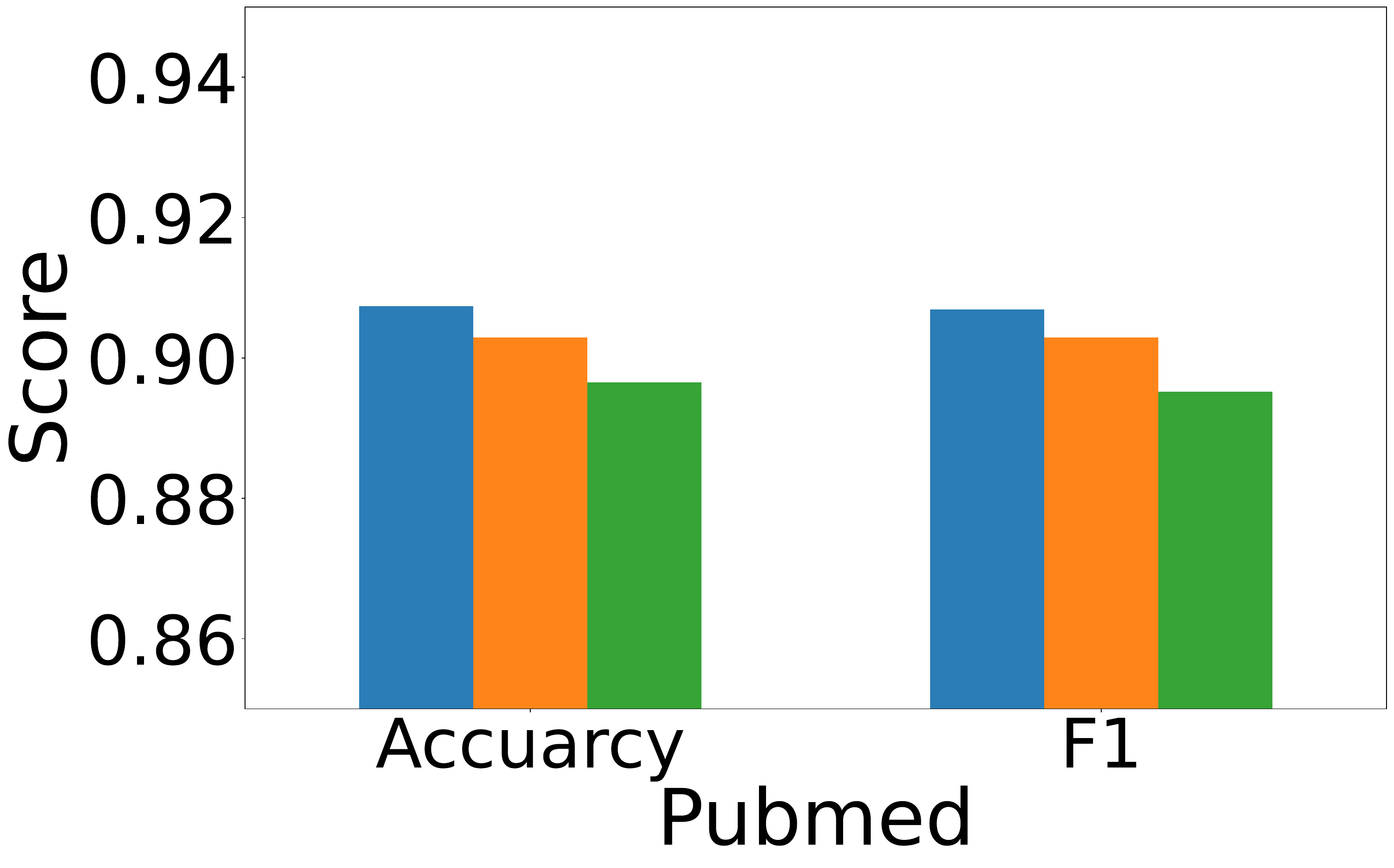}
    }
    \vskip -1.2em
    \caption{Preliminary study on the negative transfer of existing self-supervised tasks.}
    \vskip -1.5em
    \label{fig:prelinminary_study}
\end{figure}

Therefore, how to enable models to benefit from pretraining tasks in downstream tasks is an important research area. Recent studies \cite{min2023recent} in natural language processing suggest that aligning the templates of pretext tasks with those of downstream tasks enhances knowledge transfer. Inspired by this discovery, we propose to enhance graph alignment for LLMs in both self-supervised tuning and task-specific tuning. However, enhancing the alignment poses two challenges: \textbf{(i) In Stage 1}, how to design a self-supervised task that not only aligns its template with downstream tasks but also accommodates the characteristics of LLMs and graph datasets? and \textbf{(ii) In Stage 2}, how to design prompts to ensure downstream tasks are aligned with self-supervised tasks to benefit from pretexts?

In an attempt to address these challenges, we approach the problem from both two stages, aligning the task templates of each stage with those of the other. In the self-supervised tuning stage, to enhance the alignment while leveraging the strengths of LLMs and the characteristics of graph-structured datasets, a new self-supervised task should be proposed.
Since downstream tasks for graphs are typically discriminative (e.g., node classification, link prediction), the self-supervised task should be designed in a similar discriminative manner. Also, LLMs exhibit strong ability in natural language processing, and in text-attributed graphs, each node is attributed with texts, these characteristics inspire us to design a task focused on helping LLMs understand the textual attributes of nodes. 
Therefore, based on these considerations, we propose the self-supervised task, i.e., \textit{text matching}. In detail, the text matching task trains the LLM to select the most relevant texts from a set of candidates based on the given graph-structured data. Through this task, models comprehend graph-structured data in a discriminate manner, potentially enhancing its performance on downstream tasks.

In the task-specific tuning stage, we introduce category prompt methods to further align with the self-supervised task template and provide additional information to help the LLM understand the task. Specifically, we offer explanatory text for each candidate category, concatenated with its corresponding category. By transforming candidate information from a word level to a sentence level, the templates of downstream tasks are better aligned with the pretexts.
We propose two methods: \textit{category manual prompt} and \textit{category soft prompt}. In the category manual prompt method, the explanatory text is generated either manually or by a powerful LLM (e.g., GPTs \cite{achiam2023gpt}). This externally generated text helps distill external knowledge into the models. In the category soft prompt method, we replace the manually constructed explanatory text with virtual soft prompt tokens, whose embeddings can be further automatically updated through backpropagation. This allows the LLM to learn the content of the category prompts by itself, avoiding the redundant process of manual prompt design.
During the two-stage tuning process, we employ a lightweight fine-tuning approach where we freeze the parameters of LLMs and only fine-tune a few parameters (i.e., MLP projectors and several soft prompt token embeddings). By achieving alignment between the two stages, we boost the performance on downstream tasks.
Our contributions are:
\begin{itemize}[leftmargin=*]
    \item We point out that existing graph-to-token approaches for LLMs suffer from the misalignment between self-supervised training and task-specific tuning; 
    \item We propose a novel framework, GALLM, which aligns task templates between self-supervised tuning and task-specific tuning, enabling models to benefit from pretexts; and
    \item Extensive experiments on supervised learning, generalizability, and particularly zero-shot capability demonstrate GALLM's effectiveness and its potential as a graph foundational model.
\end{itemize}

\section{Related Work}

In this section, we will review related works about learning on graphs and Large Language Models for Graphs.

\subsection{Learning on Graphs}

Graph-structured data is prevalent in many aspects, such as social recommendations \cite{yang2021consisrec, luo2024spectral}, molecular structures \cite{wieder2020compact, wang2022molecular}, and intelligent transportation \cite{rahmani2023graph}. However, effectively analyzing and modeling graphs poses significant challenges due to their inherent irregularity and diverse connectivity patterns.
Graph Neural Networks (GNNs) \cite{kipf2016semi, defferrard2016convolutional, velivckovic2017graph} have emerged as a powerful framework for learning from graph data, which enable information propagation and feature extraction through iterative message passing among neighboring nodes. For example, GCN \cite{kipf2016semi} applies a convolutional operation to nodes and their neighbors, enabling the model to learn from local neighborhoods. Furthermore, GAT \cite{velivckovic2017graph} weighs the importance of neighboring nodes dynamically, focusing on relevant information. Typically, GNNs operate under a supervised setting, where models are trained for a specific task on the input graph and make inferences on the same graph. However, the difficulty of obtaining labeled data \cite{chen2024llaga} limits the performance. To address this issue, some research \cite{jin2020self, you2020graph, xia2022simgrace, zhu2021graph, hou2023graphmae2, xia2023automated} has explored self-supervised learning on graph data. For example, GraphCL \cite{you2020graph} and GCA \cite{zhu2021graph} employ contrastive learning as the self-supervised tasks, while GraphMAE \cite{hou2023graphmae2} and AutoCF \cite{xia2023automated} leverage generative tasks. Though they have made progress, their reliance on simple GNN architectures limits the models' generalizability and versatility.

\subsection{Large Language Models for Graphs}

In recent years, LLMs \cite{achiam2023gpt,touvron2023llama, perez2022discovering} have gained widespread attention for their remarkable capabilities in various NLP tasks. Furthermore, LLMs have expanded their applications beyond NLP \cite{wu2023multimodal,fathullah2024prompting,bao2023tallrec}. This emergent ability has inspired researchers to explore the use of LLMs in graph-structured data learning \cite{liu2023towards}. One significant branch of research in this area is the LLM-as-predictor approach, which utilizes LLMs as the core and sole backbone for graph learning. These methods convert graph-structured data into formats compatible with LLMs and input them alongside natural language prompts. The key challenge in employing LLMs for graph data lies in aligning the graph data with formats that LLMs can comprehend. Based on the alignment of graph data, these LLM-as-predictor methods can be categorized into two groups: graph-to-text \cite{liu2023evaluating, wang2024can, zhao2023graphtext, guo2023gpt4graph, chen2024exploring, fatemi2023talk} and graph-to-token \cite{ye2023natural,tang2024graphgpt,chen2024llaga} approaches.

Graph-to-text methods employ natural language to describe graph data. For instance, LLMtoGraph \cite{liu2023evaluating} and NLGraph \cite{wang2024can} utilize node or edge lists as input formats for graphs. GraphText \cite{zhao2023graphtext} bridges the gap between graphs and texts by translating graph structures into text with a graph-syntax tree. Moreover, GPT4Graph \cite{guo2023gpt4graph} and ``Talk like a Graph'' \cite{fatemi2023talk} introduce different graph description templates to transform graphs into languages. However, these methods are limited by the manual setup of templates to describe graphs, leading to redundancy, incomplete representation, and imprecise characterization of graphs' intrinsic features.

Contrastingly, graph-to-token methods attempt to convert graph information into sequences of tokens, which are then concatenated with the embeddings of natural language prompts and input into LLMs. To align graph tokens with text tokens, these methods utilize a projector to map graph tokens into text space through proposed various instruction tuning tasks. For example, InstuctGLM \cite{ye2023natural} introduces self-supervised link prediction tasks to enhance downstream node classification. GraphGPT \cite{tang2024graphgpt} employs a two-stage instruction tuning approach: in Stage 1, it introduces a graph matching task designed to align the order of graph tokens with text tokens; in Stage 2, it applies task-specific tuning to improve performance on downstream node classification and link prediction tasks. LLaGA \cite{chen2024llaga} proposes three different tasks (i.e., node classification, link prediction, and node description) for mixed instruction tuning. 

Although the tasks proposed in these studies have made some progress, they overly rely on supervised instruction tuning and struggle to leverage self-supervised tuning effectively, resulting in limited performance. Specifically, the self-supervised tasks introduced by previous methods differ significantly from downstream tasks, making it challenging for models to benefit from self-supervised tuning, and potentially leading to negative transfer. To address this issue, our model aligns the task templates between the two stages, enhancing performance on downstream tasks through the advantages of self-supervised tasks.

\section{Methodology}
In this section, we detail our proposed framework, illustrated in Figure \ref{fig:framework}. We utilize instruction tuning to fine-tune LLMs, employing a tuned graph projector to map graph embeddings into the textual space through various tasks (upper left of Figure \ref{fig:framework}). Our tuning process follows a two-stage paradigm: self-supervised tuning in Stage 1 and task-specific tuning in Stage 2. In Stage 1, we introduce the text matching task aligned with downstream tasks (bottom left of Figure \ref{fig:framework}). In Stage 2, we present two prompt tuning methods (i.e., manual prompts and soft prompts) to further align the task template and enhance category information (right side of Figure \ref{fig:framework}).

\subsection{Notations}
\label{sec:preliminaey}
A graph-structured data consists of a number of entities and the connecting relationships between them, where entities can be modeled as nodes in the graph, and connecting relationships can be modeled as edges. Formally, given a graph-structured data $\mathcal{G} = {\{\mathcal{V}, \mathbf{X}, {\mathcal{E}}}\}$,  $\mathcal{V}$ is the set of nodes $ {\{v_1,...,v_N\}}$ ,  and $\mathcal{E}=\{e_{ij}\}$ is the set of undirected edges. 
In this paper, we focus on the text-attributed graph, where nodes are attributed with raw textual contents $\mathbf{C} = \{\mathbf{c}_i | 1 \leq i \leq N\}$, with $\mathbf{c}_i$ being the text associated with node $v_i$. These textual attributes can be further encoded into $d-$dimensional embeddings $\mathcal{\mathbf{X}}\in \mathbb{R}^{N\times d}$ using pre-trained language models.

When conducting node-level tasks with LLMs, e.g., node classification and link prediction, we usually sample subgraphs centered at nodes as input to LLMs to (i) keep enough graph structure information of target nodes; and (ii) avoid redundant information and large computational cost for handling the whole graph. Formally, we denote $\mathcal{G}_{sub}^{h}(v) = (\mathcal{V}_v^{h}, \mathcal{E}_v^{h}, \mathbf{X}_v^{h})$ as the subgraph around node $v$, consisting of $h$-hop neighbor nodes of $v$ and all interconnection edges. We also denote $\mathcal{N}^{h}(v) = \{{v^{\prime}|v^{\prime} \in \mathcal{V}_v^{h}, v^{\prime} \neq v}\}$ as the set of $h$-hop neighbor nodes in $\mathcal{G}_{sub}^{h}(v)$. With these notations, we will sequentially introduce our instruction tuning pipeline.

\subsection{Instruction Tuning Pipeline Design}
\label{sec:pipleline_design}
\begin{figure*}
    \centering
    \includegraphics[scale=0.364]{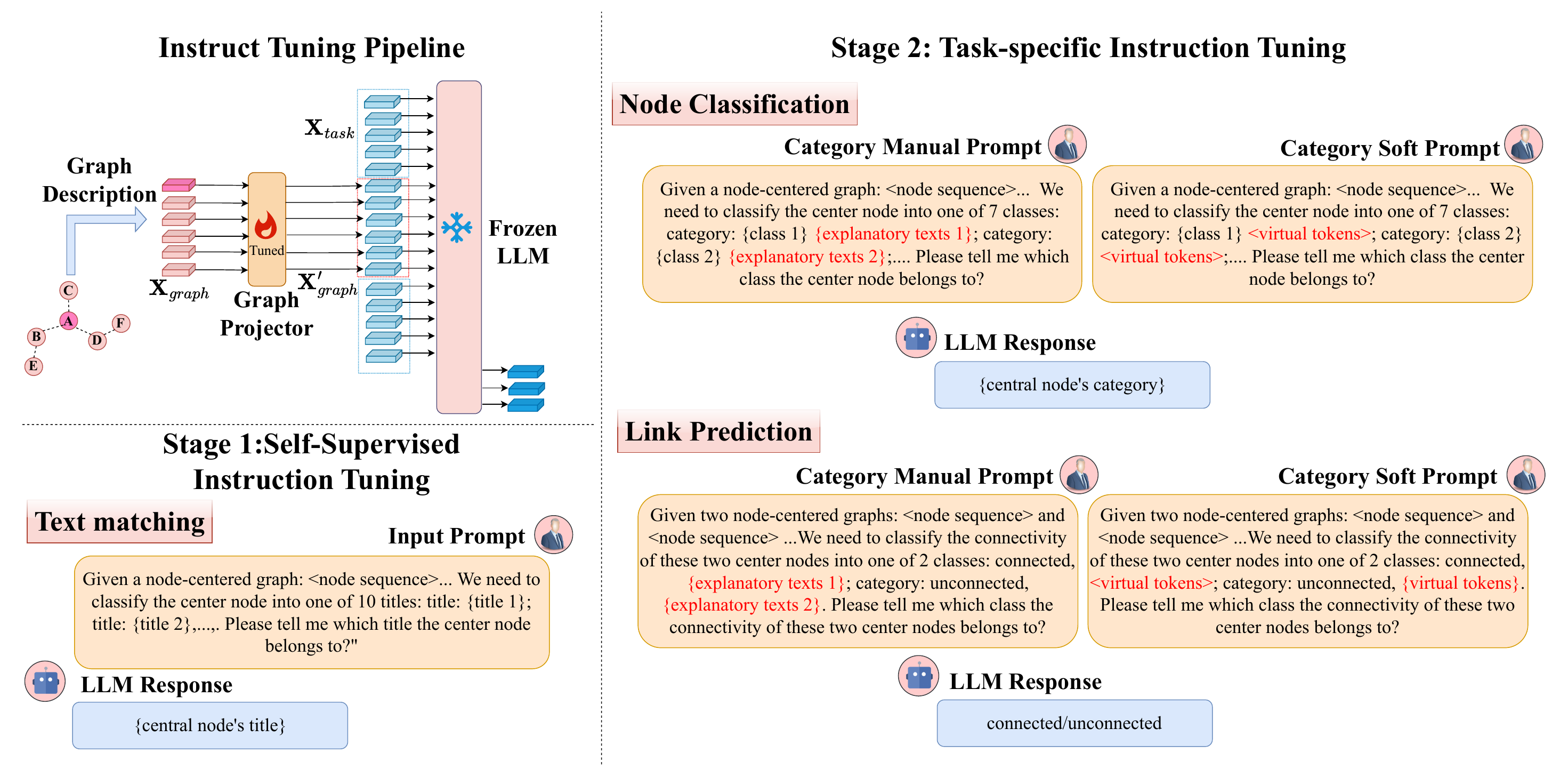}
    \vskip -2em
    \caption{The overall architecture of GALLM with two-stage instruction tuning paradigm.}
    \label{fig:framework}
    \vskip -0.5em
\end{figure*}
Instruction tuning is an effective method for incorporating additional knowledge into LLMs when applied to specific domains \cite{zhang2023instruction}. In this work, we employ a two-stage instruction tuning process. To comprehensively represent graph structure information and ensure adaptability to various tasks, we design a uniform instruction tuning pipeline, which is shown in the upper left part in Figure \ref{fig:framework}. In detail, we utilize generative LLMs as backbones which follow the auto-regressive paradigm and generate the next token as:
\begin{equation}
    P_{\theta}(\mathbf{Y}_j|\mathbf{X},\mathbf{Y}_{<j}) = {LLM}_{\theta}(\mathbf{X},\mathbf{Y}_{<j}),
\end{equation}
where $\mathbf{X}$ is the input sequence, and $LLM(\cdot)$ is the Large Language Model we utilize. $\mathbf{Y}$ is the target sequence, $\mathbf{Y}_{<j}$ is the token sequence before token $j$ and $\mathbf{Y}_j$ is the target token $j$. With all the target sequence tokens, the training loss can be formulated as:
\begin{gather}
    \mathcal{L}_{\theta} = -\sum_{j=1}^{|\mathbf{Y}|} \log P_{\theta}(\mathbf{Y}_j|\mathbf{X},\mathbf{Y}_{<j}).
\end{gather}
The core of instruction tuning lies in the design of $\mathbf{X}$ and $\mathbf{Y}$. As shown in Figure \ref{fig:soft_prompt}, we further subdivide $\mathbf{X}$ into two components: graph information $\mathbf{X}_{graph}$ and task-specific instructions $\mathbf{X}_{task}$. Given that LLMs are trained on linguistic data, the graph information should be translated into a format that can be understood by LLMs. We formulate the graph translation procedure as follows:
\begin{equation}
    \mathbf{X}_{graph} = \mathcal{T}(\mathcal{G}_{sub}^{h}(v)) = \mathcal{T}(\mathcal{V}_v^{h}, \mathcal{E}_v^{h}, \mathbf{X}_v^{h}),
\end{equation}
where $\mathcal{T(\cdot)}$ is the graph description template, which turns the graph-structured data into sequences of tokens. Previous work \cite{ye2023natural, tang2024graphgpt, chen2024llaga} has explored different description templates for translating graph information and here we use the Neighbor Detail Template \cite{chen2024llaga} since it provides an in-depth view of the central node and its surroundings. We also evaluate the compatibility with other graph description templates in Section \ref{sec:compat}.

Specifically, in the Neighbor Detail Template, for a given node $v$, we first construct a fixed-size, sampled computational tree centered around $v$ to represent the subgraph. In this tree, node $v$ serves as the root, with its 1-hop neighbors as its children, and the 2-hop neighbors as the children of the 1-hop neighbors, and so forth. For each node, a fixed number of neighbors is considered. For nodes with more neighbors than the specified size, we sample their neighbors; for nodes with fewer neighbors, we supplement them with placeholder nodes. We then conduct a level-order traversal of the computational tree, transforming the subgraph into a fixed-length node sequence. In this way, we translate the graph structure into a node sequence. 
For example, for the subgraphs in Figure \ref{fig:framework}, if we set the fixed sample number to 3, the node sequence $\mathbf{X}_{graph} = \{\mathbf{x}_A, \mathbf{x}_B, \mathbf{x}_C, \mathbf{x}_D, \mathbf{x}_E, \mathbf{x}_A, \mathbf{x}_{pad}, \mathbf{x}_A, \mathbf{x}_{pad}, \mathbf{x}_{pad}, \mathbf{x}_A, \mathbf{x}_F, \mathbf{x}_{pad}\}$, where $\mathbf{x}_{pad}$ denotes embedding of the pad token and is set to a zero vector.

After obtaining the node sequences, we use a projector to map the graph embedding to the text embedding space, thus allowing LLMs to understand the graph information:
\begin{equation}
    \mathbf{X}_{graph}^{\prime} = Projector(\mathbf{X}_{graph}).
\end{equation}
Typically, we use a linear layer as the projector. With the projected graph embeddings, we concatenate them with the text embeddings of task-specific instruction to form the input sequence for the LLM:
\begin{equation}
    \mathbf{X} = Concatenate(\mathbf{X}_{task}, \mathbf{X}_{graph}^{\prime}).
\end{equation}
Here $\mathbf{x}_{task}$ denotes the embeddings of the instruction texts encoded by the LLM's embedding layer. In detail, when inputting instruct texts into the LLM, we use placeholder tokens to fill the positions of the graph tokens. After obtaining the text embeddings $\mathbf{X}_{task}$, we replace the embeddings of these placeholder tokens with the graph embeddings $\mathbf{X}_{graph}$. The design of effective instruction tuning tasks hinges on definitions of task-specific instructions $\mathbf{X}_{task}$ and target sequences $\mathbf{Y}$. We propose a two-stage tuning process that follows the uniform pipeline, differing only in $\mathbf{X}_{task}$ and $\mathbf{Y}$. Specifically, we introduce the text matching task for Stage 1 in Section \ref{sec:stage1} and the category prompt methods for Stage 2 in Section \ref{sec:stage2}.

To streamline the tuning process, we use a lightweight strategy. During instruction tuning, we freeze the LLM backbone parameters and optimize only a few. Specifically, in both Stage 1 (i.e., Section \ref{sec:stage1}) and Stage 2 of the category manual prompt approach (i.e., Section \ref{sec:mp}), we fine-tune only the graph projector parameters. In Stage 2 of the category soft prompt approach (i.e., Section \ref{sec:sp}), we fine-tune the graph projector, soft prompt embeddings, and soft prompt encoder. This minimal parameter tuning compared to the LLM backbone significantly enhances efficiency. The following sections detail our proposed tasks for both two stages.

\subsection{Self-Supervised Instruction Tuning}
\label{sec:stage1}
Self-supervised fine-tuning leverages extensive unlabeled data, enabling LLMs to acquire knowledge of graph structures, thereby enhancing the model's performance on downstream tasks. However, the design of self-supervised tasks involves two critical considerations: (\textbf{i}) \textit{the template of self-supervised tasks should be aligned with that of downstream tasks to facilitate knowledge transfer}, and (\textbf{ii}) \textit{the self-supervised tasks must be compatible with the characteristics of LLMs and text-attributed graph datasets}. 
For the first consideration, as downstream tasks for graphs are typically discriminative (e.g., node classification, link prediction), self-supervised tasks should also be designed in a similar discriminative manner. For the second consideration, LLMs exhibit a strong capacity for understanding texts, and in text-attributed graphs, each node is associated with textual features. This motivates us to utilize the textual attributes of nodes in designing the self-supervised task. Based on these considerations, we propose a novel self-supervised task: \textit{text matching}.
\subsubsection{Self-Supervised Task Design}
In the text matching task, we provide graph data and candidate texts to the LLM, requiring it to select the most relevant text for the input graph data. The nodes' textual attributes (e.g., paper titles in the citation network dataset) serve as these candidate texts. Among these, the positive text (i.e., ground truth) corresponds to the textual attribute of the central node; while the negative texts are derived from the textual attributes of randomly sampled nodes.
For example, as shown in the bottom left part of Figure \ref{fig:framework}, the task-specific instruction for citation network can be ``\textit{Given a node-centered graph: <node sequence>, each node represents a paper. The first token represents the central node of the subgraph and the remaining represents the neighbors. We need to classify the center node into one of the 10 titles: title: $\mathbf{c}^{pos}$; title: $\mathbf{c}^{neg}_j$; ...; title: $\mathbf{c}^{neg}_K$. Please tell me which title the center node belongs to?}'', the target sequence is the title of the central node paper $\mathbf{Y} = \mathbf{c}^{pos}$. 

The text matching task aims to select correct text attributes (e.g., titles) for each central node, thereby enhancing LLMs' understanding of graph data. Moreover, since downstream tasks are formulated in a similar discriminative manner, our model requires only a simple modification of the content to be matched during downstream tasks, changing candidate texts to candidate labels. For node classification tasks, candidate labels represent nodes' categories; while for link prediction tasks, they indicate connection presence or absence. Thus, this template alignment helps transfer knowledge from the text matching task to downstream tasks more effectively.

\subsubsection{Mixing Negative Sampling Strategy.}
Previous works \cite{robinson2020contrastive, kalantidis2020hard} have demonstrated that incorporating negative samples of varying difficulty levels can enhance the model's robustness. Inspired by this, we categorize our negative text samples into two types: easy negative samples and hard negative samples. Generally, the textual attributes of neighboring nodes tend to be quite similar. For instance, in a citation network, a paper titled ``Hard negative mixing for contrastive learning'' is cited by a paper titled ``Contrastive Learning with Hard Negative Samples'', which results in these two papers being neighboring nodes in the graph, with similar titles. Therefore, when constructing negative samples, we obtain \textit{hard negative samples} by sampling from neighboring nodes as:
\begin{equation}
    \mathcal{C}^{hard-neg} \sim Sample(\mathcal{N}^h_v).
\end{equation}
We sample \textit{easy negative samples} from non-neighboring nodes:
\begin{equation}
    \mathcal{C}^{easy-neg} \sim Sample(\mathcal{V} \setminus \mathcal{N}^h_v).
\end{equation}
The final negative sample set is $\mathcal{C}^{neg} = \{\mathcal{C}^{easy-neg}, \mathcal{C}^{hard-neg}\}$. Generally, easy negative samples help the model develop basic graph processing abilities, while hard ones further enhance these skills. Thus, we include more easy negative samples than hard ones to ensure the model maintains its fundamental abilities while improving its capacity to distinguish challenging samples.

\subsection{Task-specific Instruction Tuning}
\label{sec:stage2}
\begin{figure}
    \centering
    \includegraphics[scale=0.68]{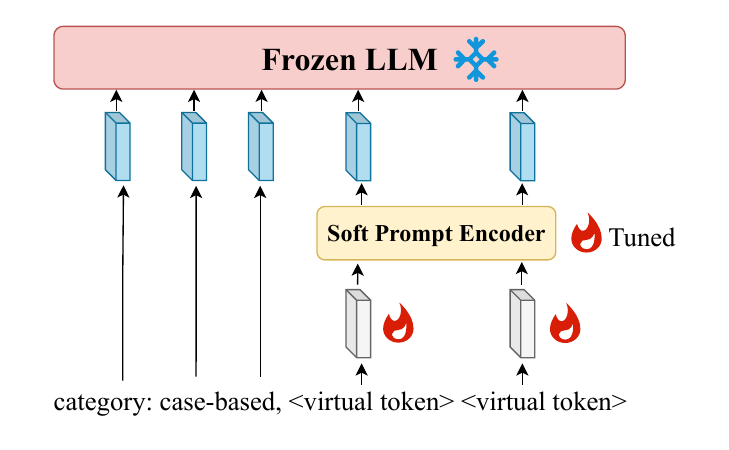}
    \vspace{-1.5em}
    \caption{Illustration of category-enhanced soft prompt tuning. We show only the category prompt, keeping other parts (e.g., graph data input format) unchanged. In this tuning, the parameters of virtual token embeddings, soft prompt encoder, and graph projector are tuned.}
    \label{fig:soft_prompt}
    \vspace{-1em}
\end{figure}

In the second stage, we propose task-specific instruction tuning, which customizes LLM's reasoning behavior for downstream tasks. Specifically, the LLMs' reasoning process can be understood as follows: first, LLMs understand the input graph information and the task instructions. Subsequently, LLMs utilize internal parameterization knowledge to perform graph tasks based on the understanding of graph information and task instructions. For instance, in node classification tasks, LLMs first comprehend the input graph and category information, and then categorize the nodes into the most appropriate classes based on their internal parameterized knowledge. However, existing methods typically provide category information at the word level (i.e., the name of categories), which limits LLM's comprehension of categories. Additionally, in self-supervised tasks, the candidate texts are provided at the sentence level. Therefore, to \textbf{enhance the LLM's understanding of category information} and further \textbf{align downstream task templates with pretexts}, we introduce the category prompt tuning method. The goal of our category prompt method is to provide explanatory content for categories, reformulating category information from word levels to sentence levels. Specifically, we propose two category prompt methods: \textit{manual prompt tuning} and \textit{soft prompt tuning}.
\subsubsection{Category-Enhanced Manual Prompt Tuning.}
\label{sec:mp} 
In category-enhanced manual prompt tuning, we construct an explanatory sentence for each category and append it following the category in the input prompt. For instance, for citation networks, the input prompt for the category ``case-based'' is: ``category: case-based, \textit{which analyzes real-world cases to draw conclusions};'' where the italicized words represent the explanatory text we added. These explanatory texts are meticulously crafted with insights from external experts. Additionally, we harness the capabilities of robust general LLMs (e.g., ChatGPT \cite{achiam2023gpt}) to automatically generate these explanations. This method effectively distills external knowledge into our model, thereby boosting its performance. 
For link prediction, we reformulate the task into a binary classification template and leverage manual prompts to enhance each category (i.e., connected and unconnected). The input prompts are depicted in the Figure \ref{fig:framework}. The complete instruction prompts and explanatory texts can be found in Appendix \ref{sec:full_prompt_appendix}.

\subsubsection{Category-Enhanced Soft Prompt Tuning.}
\label{sec:sp}
While manual prompts can incorporate external knowledge, the use of discrete manual prompts often results in unstable performance (e.g., changing a single word in the prompt can lead to a significant drop in performance) \cite{liu2023gpt, li2021prefix, lester2021power}. To address this issue, we propose an alternative approach: category-enhanced soft prompt tuning, which is illustrated in Figure \ref{fig:soft_prompt}. Soft prompt methods \cite{liu2023gpt, li2021prefix, lester2021power} use trainable continuous prompt embeddings in concatenation with discrete prompts. In our methods, we replace category manual prompt tokens with virtual soft prompt tokens. For example, the input category text can be: ``category: case-based, \textit{<soft prompt token> <soft prompt token>};''. The embeddings of these virtual prompt tokens are initialized randomly. Subsequently, a soft prompt encoder (i.e., a two-layer MLP) is utilized to further encode these embeddings. After encoding, these continuous prompt embeddings are concatenated with the discrete prompt tokens and provided as input to the LLMs. The category soft prompts are updated via backpropagation to optimize the task objective, enabling the LLM to autonomously learn the explanatory context of categories. This can be formulated as
\begin{equation}
    \mathop{\min}_{\theta_{gp},\theta_{sp}} \ \ \mathcal{L}_{\theta} = -{\sum}_{j=1}^{|\mathbf{Y}|} \log P_{\theta} (\mathbf{Y}_j|\mathbf{X},\mathbf{Y}_{<j}),
\end{equation}
where $\theta_{gp}$ denotes the parameters of the graph projector, and $\theta_{sp}$ denotes the parameters of virtual token embeddings and the soft prompt encoder. Similarly, for link prediction, we utilize virtual soft prompt tokens to enhance each category.


\section{EXPERIMENTS}

In this section, we demonstrate the effectiveness of our proposed GALLM across multiple scenarios. Through the experiments, we aim to answer the following questions: (i) \textbf{Q1:} How does our framework GALLM perform compared to baseline models in traditional supervised graph learning settings? (ii) \textbf{Q2:} What is the zero-shot ability of our model, specifically with regard to self-supervised learning and cross-dataset settings? (iii) \textbf{Q3:} What is the multi-dataset generalization ability of our model across multiple datasets?  (iv) \textbf{Q4:} How does each key module in GALLM contribute to enhancing performance? (v) \textbf{Q5:} How compatible is our GALLM with different graph descriptions and base LLMs (i.e., LLaMA 3)?

\subsection{Experiment Setup}
\label{sec:experiment}
\subsubsection{Datasets.}
\label{sec:dataset}


We choose four widely-used datasets: Cora \cite{yang2016revisiting}, Pubmed \cite{he2023explanations}, Arxiv \cite{hu2020open} and Instagram \cite{kim2020multimodal}. These datasets originate from different domains, with the first three derived from citation networks and the last from social networks. 
For citation network datasets, nodes represent academic papers and edges denote citation relationships. Each node is attributed with a title and an abstract and is categorized into distinct research topics. 
For the social network dataset Instagram, nodes represent users, and edges signify ``following'' relationships. Users are attributed with follow lists, personal introductions, and tags, and are categorized into two classes: normal users and commercial users. The statistics are presented in Tabel \ref{tab:datasets} and more details are in Appendix \ref{sec:dataset_appendix}.

\begin{table}
\caption{Statistics of experiment datasets.}
\vspace{-1em}
\scalebox{0.9}{
\setlength{\tabcolsep}{0.75 mm}{
\begin{tabular}{@{}cc|ccccc@{}}
\toprule
& Dataset     & \# Nodes & \# Edges & \# Class & Splitting & Domain \\ \hline \hline
& Cora      & 2708     & 5429   & 7  & 6:2:2 &  citation     \\ \hline
& Pubmed    &  19717     & 44338   & 3  & 6:2:2 & citation   \\ \hline
& Arxiv   &  169343   & 1166243    & 40  & 6:2:3 & citation     \\ \hline
& Instagram   &  11339   & 144010      & 2  & 1:1:8 & social network   \\ \bottomrule
\end{tabular}}}
\vspace{-2pt}
\label{tab:datasets}
\end{table}

\begin{table*}
\centering
\caption{Performance comparison on node classification under supervised settings. Here bold signifies the best result across all methods, while underline highlights the best baseline result.}
\vspace{-1em}
\scalebox{0.96}{
 \begin{tabular}{c|c|cc|cc|cc|cc}
 \toprule
 \multirow{2}{*}{Method} & Datasets    & \multicolumn{2}{c|}{Cora} & \multicolumn{2}{c|}{Pubmed} & \multicolumn{2}{c|}{Arxiv} & \multicolumn{2}{c}{Instagram}   \\ \cline{2-10}
                        & Metric      & Acc  & F1     & Acc    & F1 & Acc   & F1 & Acc &F1  \\ \hline \hline
 \multirow{10}{*}{Baselines}               
                        & GCN     & \underline{0.8930} & \underline{0.8795} & 0.8948 & 0.8904 & 0.6765 & 0.2625 & 0.6187 & 0.5524 \\
                        & GraghSage & 0.8782 & 0.8677 & \underline{0.9054} & \underline{0.9051} & 0.6991 & 0.3339  & 0.6062 & \underline{0.5552}     \\
                        & GAT & 0.8708 & 0.8632 & 0.8811 & 0.8760 & 0.6788 & 0.2648 & 0.6064 & 0.5502 \\
                        & SGC & 0.8875 & 0.8718 & 0.8991 & 0.8943 & 0.6728 & 0.2334 & 0.6220 & 0.5546 \\ 
                        \cline{2-10}
                        & UniMP & 0.8782 & 0.8632 & 0.9034 & 0.9017 & 0.6953 & 0.2977  & 0.6096 & 0.5484 \\
                        & NodeFromer  & 0.8155 & 0.7927 & 0.8910 & 0.8917 & 0.6869 & 0.5064  & 0.5778 & 0.5385   \\ 
                        \cline{2-10}
                        & GraphCL & 0.8653 & 0.8450 & 0.9039 & 0.9028 & 0.6611 & 0.2933 & 0.6384 & 0.5512  \\
                        & SimGRACE & 0.8745 & 0.8632 & 0.9049 & 0.9039 & 0.6985 & 0.3264 & \underline{0.6404} & 0.5420 \\
                        \cline{2-10}
                        & GraphGPT & 0.8392 & 0.8272 & 0.8176 & 0.8099 & 0.7159 & 0.5458 & 0.6071 & 0.5050 \\
                        & LLaGA  & 0.8632 & 0.8434 & 0.9021 &  0.9020 & \underline{0.7323} & \underline{0.5660} & 0.6133 & 0.5549 \\
                        \cline{1-10}
 \multirow{2}{*}{Ours}
                        & GALLM-sp  & 0.8854 & 0.8757 & 0.9087 & 0.9086 & \textbf{0.7391} & \textbf{0.5775} & 0.6274 & 0.5203 \\
                        & GALLM-mp & \textbf{0.8965} & \textbf{0.8934} & \textbf{0.9115} & \textbf{0.9105} & 0.7330 & 0.5664 & \textbf{0.6460} & \textbf{0.5583}
                         \\ \bottomrule
 \end{tabular}}
\label{tab:perform_nc}
\vskip -5pt
\end{table*}

\subsubsection{Baselines and Implementation Details.} 
\label{sec:baselines}
To evaluate our proposed GALLM, we compare it with various models grouped into four categories: (i) Traditional GNN models, including GCN \cite{kipf2016semi}, GAT \cite{velivckovic2017graph}, GraphSage \cite{hamilton2017inductive}, and SGC \cite{wu2019simplifying}; (ii) Transformer-based models, including UniMP \cite{shi2020masked} and NodeFormer \cite{wu2022nodeformer}; (iii) Self-supervised GNNs, including GraphCL \cite{you2020graph} and SimGRACE \cite{xia2022simgrace}; (iv) State-of-the-art LLMs for graphs, including GraphGPT \cite{tang2024graphgpt} and LLaGA \cite{chen2024llaga}. For zero-shot ability assessment, we further include ZeroG \cite{li2024zerog}, which focuses on cross-dataset zero-shot performance. 
We exclude methods like TAPE \cite{he2023harnessing} and LLM-GNN \cite{chen2023label} since they belong to a different branch. These methods are highly task-specific, following a per-model-per-dataset manner and relying on additional explanations, whereas we aim to develop a graph foundation model that operates across multiple scenarios.

For GALLM, we adopt Vicuna-7B-v1.5-16K as our foundation base model and utilize a pre-trained Sentence-BERT \cite{reimers2019sentence} to encode nodes' raw text features. More baseline descriptions and implementation details are in Appendix \ref{sec:appendix_baseline} and Appendix \ref{sec:imple_datail_appendix}.


\subsubsection{Evaluation Settings.} 
\label{sec:setting}
For node classification, we follow prior works \cite{hu2020open, tang2024graphgpt, chen2024llaga} and randomly split the datasets into training, validation, and test sets with proportions of 6:2:2 for Cora and PubMed, and 6:2:3 for Arxiv. For the Instagram dataset, where node classification resembles anomaly detection typically conducted in a few-shot setting, we follow prior work \cite{huang2024can} and apply a split ratio of 1:1:8.
For link prediction, consistent with previous work \cite{chen2024llaga}, we randomly select node pairs from the node classification training set for training and from the test set for evaluation, ensuring the edge-level training sets match the size of the node-level training sets. To evaluate the performance of our model, we adopt two widely used metrics, i.e., accuracy (Acc) and macro-F1 (F1).

\subsection{Overall Performance Comparison}
\label{sec:pc}

\begin{table}
\centering
\caption{Performance comparison on link prediction under supervised settings.}
\vspace{-1em}
\scalebox{0.9}{
 \begin{tabular}{c|c|cc|cc}
 \toprule
 \multirow{2}{*}{Method} & Datasets  & \multicolumn{2}{c}{Cora} & \multicolumn{2}{c}{Pubmed}  \\ \cline{2-6} 
 & Metric & Acc & F1 & Acc & F1 \\ \hline \hline
 \multirow{2}{*}{Baselines}               
                        & GraphGPT & 0.5985 & 0.5985 & 0.7360 & 0.7339  \\
                        & LLaGA  & 0.6156 & 0.5990 & 0.7241 & 0.7239 \\
                        \hline 
 \multirow{2}{*}{Ours}  & GALLM-sp & \textbf{0.6230}  & \textbf{0.6154} & 0.7321 & 0.7319                              \\
                        & GALLM-mp & 0.6156  & 0.5975 & \textbf{0.7373} & \textbf{0.7373}
                         \\ \bottomrule
 \end{tabular}}
\label{tab:perform_lp}
\vskip -10pt
\end{table}

To answer Q1, we evaluate state-of-the-art models and our proposed GALLM in a supervised setting, where models are trained and tested on the training and test set of the same dataset, respectively.
\subsubsection{Performance on Node Classification.}
We mainly focus on the node classification task, and the results of our model and baselines are presented in Table \ref{tab:perform_nc}. In the table, GALLM-sp denotes our model with category-enhanced soft prompt tuning, and GALLM-mp denotes our model with category-enhanced manual prompt tuning. Based on these results, we can draw the following observations: 
\begin{itemize}[leftmargin=*]
    \item Overall, GALLM achieves state-of-the-art performance, exceeding baselines across all metrics and datasets, which highlights its strength in traditional supervised settings.
    \item Both the category soft prompt and manual prompt methods demonstrate distinct advantages. When labeled data is limited (e.g., Cora, Pubmed, and Instagram), category manual prompts excel by incorporating external knowledge and avoiding the introduction of additional parameters that require fine-tuning. Conversely, in cases of abundant labeled data (e.g., Arxiv), category soft prompts show superior performance. This enhancement is attributed to the model's ability to update the parameters of the soft prompt embeddings by itself, allowing the category prompts to be more effectively tailored to downstream tasks.
    \item LLMs show great potential for tackling complex graph tasks. The Arxiv dataset poses a significant challenge for node classification due to its many node categories. In this dataset, LLM-based methods markedly outperform other approaches, showcasing their emergent abilities in handling complex graph tasks.
\end{itemize}




\subsubsection{Performance on Link Prediction.}
Additionally, we also conduct evaluations on the link prediction task. Since LLMs output discrete responses rather than continuous values, we choose LLM-based methods as baselines for a fair comparison, and the results are shown in Table \ref{tab:perform_lp}. The results show that GALLM also surpasses other LLM-based models in link prediction, likely due to aligned task templates that enable models to benefit from pretexts.

\subsection{Zero-Shot Ability}

To answer Q2, we explore our model's zero-shot ability under two settings: the self-supervised setting and the cross-dataset setting.

\subsubsection{Self-Supervised Setting.} 
In the self-supervised setting, the model is trained solely through self-supervised learning (for models with such tasks) and then tested on the same dataset, without using labeled data. We investigate model performances on the Cora, Pubmed, and Arxiv datasets. For self-supervised tasks, GraphGPT employs graph matching, LLaga utilizes node descriptions, and our GALLM is tuned with the text matching task. ZeroG is evaluated directly, as it lacks self-supervised tasks. 

Experimental results in Table \ref{tab:zero-shot} lead to several conclusions: (i) In the self-supervised setting, our model demonstrates remarkable performance. This success is attributed to our self-supervised tasks, which leverage the characteristics of LLMs and graph datasets while aligning with downstream task templates. In this way, the knowledge learned in the self-supervised stage can effectively transfer to downstream tasks, showcasing strong capabilities even without supervised tuning. (ii) In contrast, though LLaGA and GraphGPT are also trained on self-supervised tasks, their task templates and training objectives differ significantly from downstream tasks, thereby restricting the effective application of acquired knowledge.

\subsubsection{cross-dataset Setting.}   In the cross-dataset setting, the model learns from a source dataset (including self-supervised learning for models with such tasks) and is then tested on a target dataset, without any access to the target dataset during training. We choose Arxiv as the source dataset with Cora and Instagram as the target dataset. Experimental results are shown in Table \ref{tab:zero-shot}. We can observe that our model also exhibits strong knowledge transfer capabilities. By aligning task templates, the knowledge learned by LLMs can be seamlessly transferred across datasets. Hence, in cross-dataset scenarios, our model demonstrates strong generalization capabilities.

\subsection{Generalizablity of GALLM}


\begin{table}
\caption{Performance of zero-shot ability.}
\vspace{-1em}
\scalebox{0.9}{
\setlength{\tabcolsep}{0.75 mm}{
\begin{tabular}{@{}cc|ccc|ccc@{}}
\toprule
& Settings    & \multicolumn{3}{c|}{Self-supervised tuning} & \multicolumn{2}{c}{Cross-dataset} \\ \hline
& Datasets     & Cora & Pubmed & Arxiv & Arxiv-Cora & Arxiv-Instagram \\ \hline \hline
& GraphGPT      & 0.0776     & 0.3054   & 0.0113  & 0.0111 & 0.1659     \\ \hline
& LLaGA    & 0.2989      & 0.4108   & 0.0059 & 0.2514 & 0.1750 \\ \hline
& ZeroG    & 0.4077      &0.3747    & 0.0503 & 0.2232 & 0.1741 \\ \hline
& GALLM-mp   & \textbf{0.5804}    & \textbf{0.4791}      & \textbf{0.1309}  & \textbf{0.2957}  & \textbf{0.1756}    \\ \bottomrule
\end{tabular}}}
\label{tab:zero-shot}
\vskip -10pt
\end{table}
To address Q3, we explore the generalizability of our model in this section, specifically investigating its ability to handle multiple datasets. In our experiments, models are first trained on node classification tasks using both the Pubmed and Arxiv datasets and then evaluated on each dataset. The results are presented in Table \ref{tab:gen}.

The results indicate that GALLM demonstrates robustness when trained on multiple datasets. Training a model simultaneously on many different datasets can lead to catastrophic forgetting \cite{kirkpatrick2017overcoming}, resulting in performance decline. Traditional GNN models often experience performance degradation during joint training across multiple datasets. In contrast, GraphCL maintains its performance through self-supervised tasks that effectively model graph structures. Also, our GALLM, when trained across different datasets, still achieves superior performance, which is because our approach enables the LLM to accurately learn the graph structure.


\begin{table}
\centering
\caption{Performance comparison on node classification for generalizability investigation.}
\vspace{-1em}
\scalebox{0.9}{
 \begin{tabular}{c|c|cc|cc}
 \toprule
 \multirow{3}{*}{Method} & Training set  & \multicolumn{4}{c}{Pubmed+Arxiv}  \\ \cline{2-6}
 & Test set    & \multicolumn{2}{c|}{Pubmed} & \multicolumn{2}{c}{Arxiv}      \\ \cline{2-6}
 & Metric & Acc & F1 & Acc & F1 \\ \hline \hline
 \multirow{10}{*}{Baselines}               
                        & GCN     & 0.8953 & 0.8918  & 0.6640 & 0.2315   \\
                        & GraghSage & 0.8981  & 0.8965  & 0.6936 & 0.3054  \\
                        & GAT & 0.8821  & 0.8778  & 0.6626 & 0.2318 \\
                        & SGC & 0.8864 & 0.8825  & 0.6465 & 0.1963 \\ 
                        \cline{2-6}
                        & UniMP & 0.9021 & 0.9019 & 0.6977 & 0.3015 \\
                        & NodeFromer  & 0.8846 & 0.8859  & 0.6510 & 0.2335   \\ 
                        \cline{2-6}
                        & GraphCL & 0.9021  & 0.9006 & 0.6758 & 0.3227 \\
                        & SimGRACE & 0.8993  & 0.8970  & 0.6730 & 0.3145 \\
                        \cline{2-6}
                        & GraphGPT & 0.7964 & 0.7911 & 0.7116 & \underline{0.4200}  \\
                        & LLaGA  & \underline{0.9031} & \underline{0.9024} & \underline{0.7281} & 0.3164 \\
                        \hline 
 \multirow{1}{*}{Ours}   & GALLM-mp & \textbf{0.9120}  & \textbf{0.9114} & \textbf{0.7306} & \textbf{0.5603}
                         \\ \bottomrule
 \end{tabular}}
\label{tab:gen}
\vskip -0.5em
\end{table}

\subsection{Ablation Study}
\label{sec:as}

\begin{table}
\small
\caption{The ablation study performance.}
\vspace{-1em}
\scalebox{0.83}{
{
\begin{tabular}{@{}cc|cc|cc|cc|cc@{}}
\toprule
&Dataset & \multicolumn{2}{c}{Cora}  & \multicolumn{2}{c}{Pubmed}  & \multicolumn{2}{c}{Arxiv}  & \multicolumn{2}{c}{Instagram}  \\ \hline
&Metric   & Acc      & F1    & Acc     & F1      & Acc    & F1    & Acc      & F1         \\ \hline  \hline
&GALLM-mp  & 0.8965 & 0.8934 & 0.9115 & 0.9105 & 0.7330 & 0.5664 & 0.6460 & 0.5583\\ 
&\textit{w/o} tm    & 0.8724 & 0.8604 & 0.9076 & 0.9061 & 0.7285 & 0.5587 & 0.6405 & 0.5580     \\ 
\cline{1-10}
&GALLM-sp  & 0.8854 & 0.8757 &  0.9087 & 0.9086 & 0.7391 & 0.5775 & 0.6274 & 0.5203  \\ 
&\textit{w/o} tm    & 0.8743 & 0.8588 & 0.8988  & 0.8983 & 0.7330 & 0.5664 & 0.6215 & 0.5187        \\ 
\cline{1-10}
&\textit{w/o} cp    & 0.8780  & 0.8667  & 0.9031 & 0.9025 & 0.7310 & 0.5632 & 0.5982 & 0.5021        \\ 
\bottomrule
\end{tabular}}}
\label{tab:ablation}
\vspace{-12pt}
\end{table}
To answer Q4, we verify the importance of the key modules, \textit{i.e.}, the self-supervised text matching tuning, and the category prompt tuning. For both GALLM-sp and GALLM-mp, we construct slim versions, i.e., GALLM-sp \textit{w/o} tm and GALLM-mp \textit{w/o} tm, which are tuned solely on downstream tasks without self-supervised tasks. Also, we construct GALLM \textit{w/o} cp, where both category manual prompt and category-enhanced soft prompt tuning are removed, yet the models are still fine-tuned with two-stage tasks. The results in Table \ref{tab:ablation} demonstrate the effectiveness of these modules.

From Table \ref{tab:ablation}, we can observe that (i) self-supervised text matching task is effective for boosting model performance. Compared to the slim version (i.e., both GALLM-sp \textit{w/o} sp and GALLM-mp \textit{w/o} mp), the complete GALLM (i.e., both GALLM-sp and GALLM-mp) achieves better performance. This improvement is due to the alignment with downstream tasks and adaptation to the characteristics of LLMs and graph datasets. In this way, the knowledge learned during self-supervised tasks can be effectively transferred to downstream tasks. (ii) Category prompt tuning is also crucial for graph learning. Compared to the slim version GALLM \textit{w/o} cp, the complete GALLM shows enhanced performance. This enhancement is attributed to further alignment with pretexts and the explanatory text provided to LLMs, which not only deepens the LLM's understanding of categories but also facilitates knowledge transfer.



\subsection{Compatibility Investigation}
\label{sec:compat}
\begin{figure}[t]
    \centering
    \subfigure[Model performance with Text-Graph Grounding Pre-trained GNNs]{
    \includegraphics[scale=0.08]{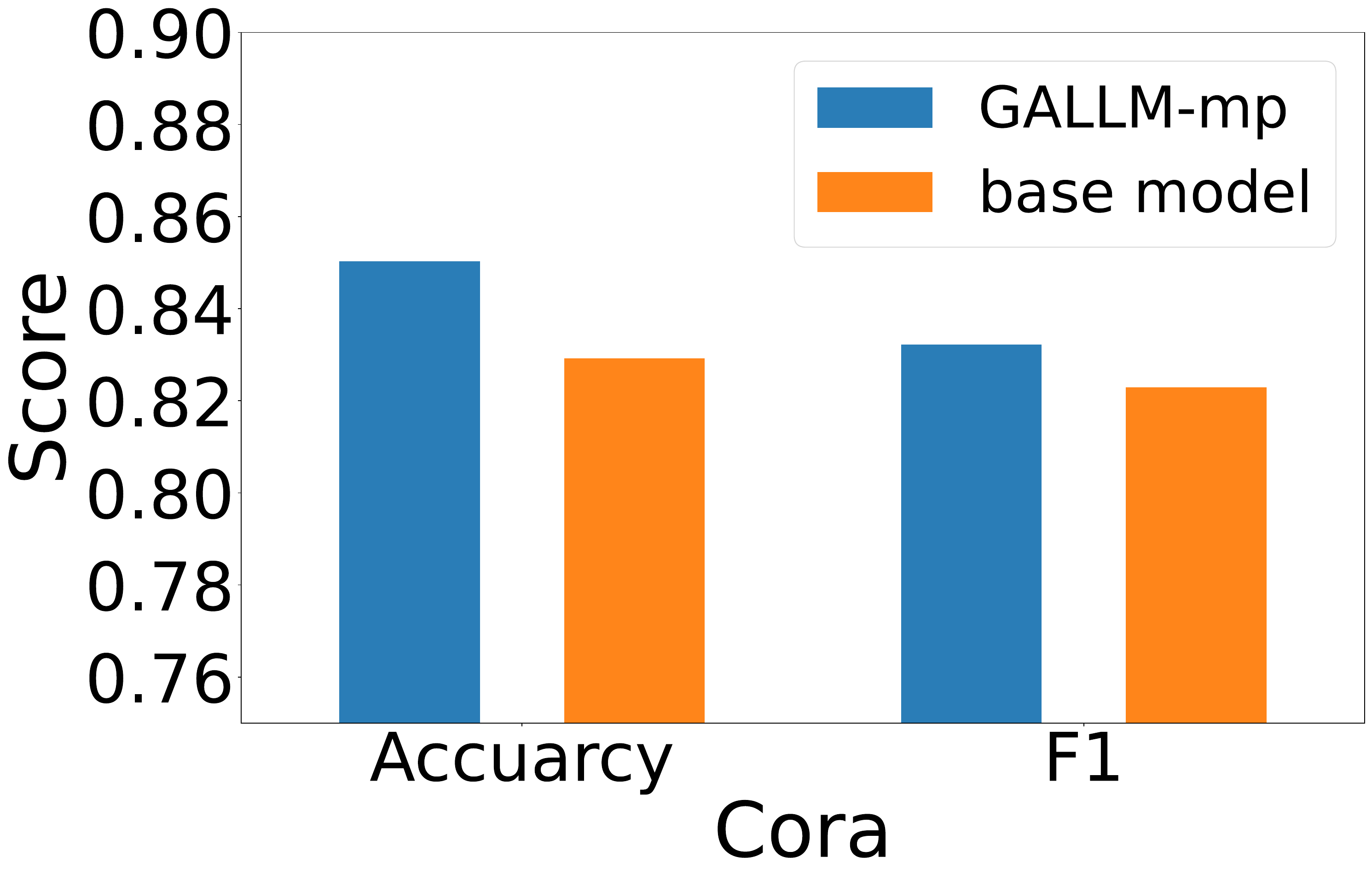}
    \includegraphics[scale=0.08]{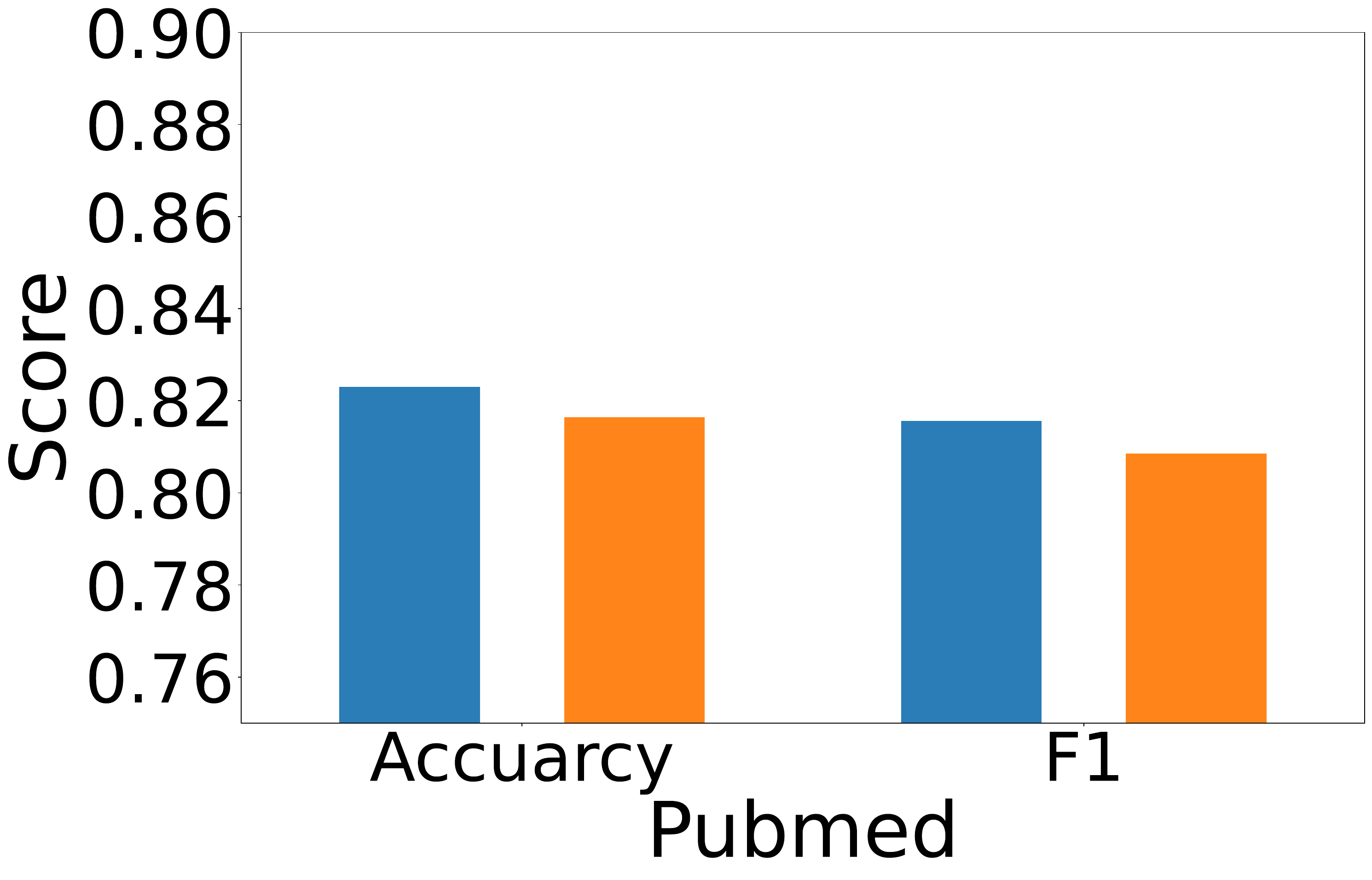}
    }
    \vskip -1pt
    \subfigure[Model performance with LLaMA3-8b]{
    \includegraphics[scale=0.08]{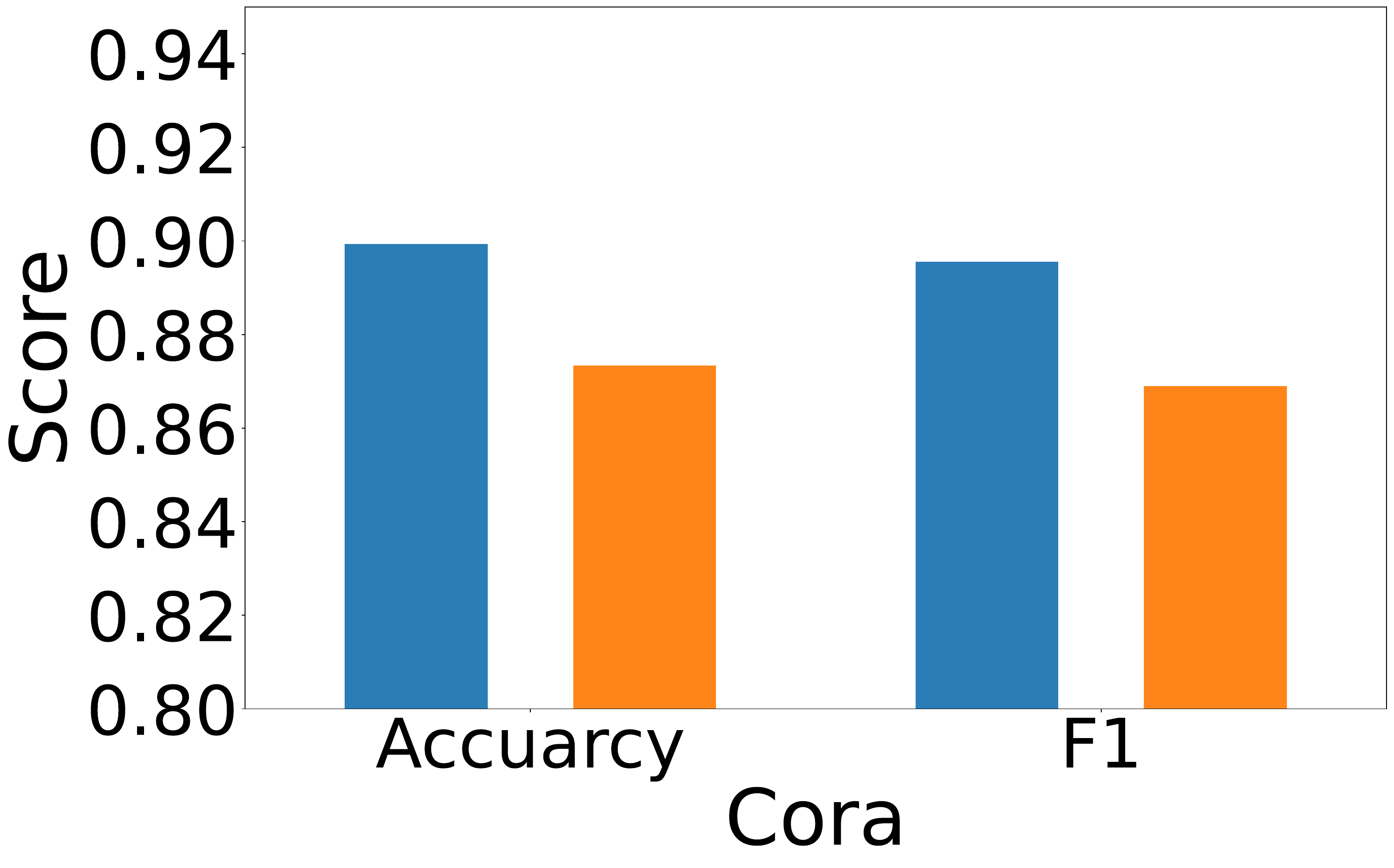}
    \includegraphics[scale=0.08]{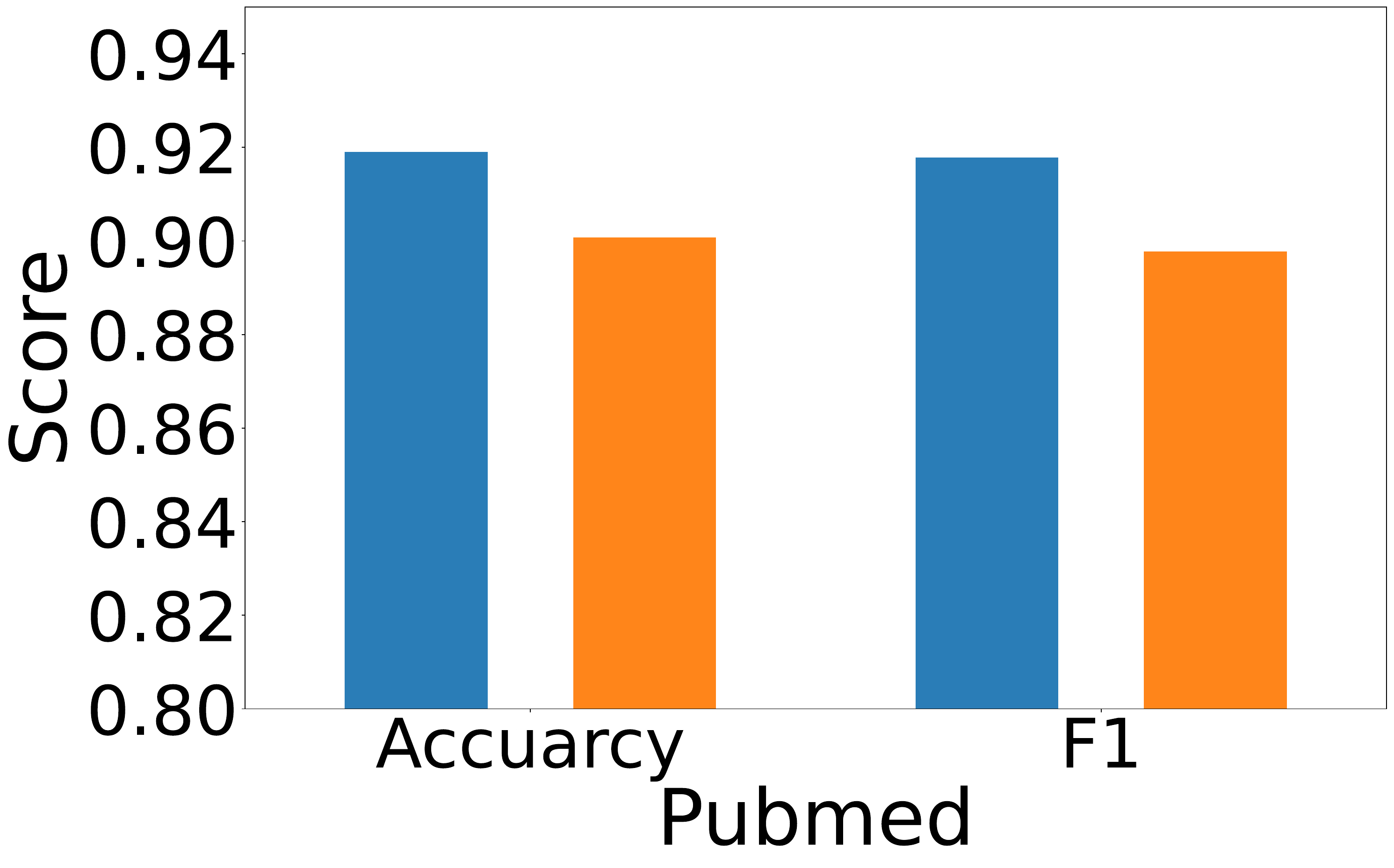}
    }
    \vskip -1.5em
    \caption{Compatibility evaluation with different graph description templates and base LLM models.}
    \vskip -1em
    \label{fig:compatibility}

\end{figure}

To answer Q5, we explore the compatibility in this section, specifically focusing on different graph description templates and base LLM models. In our initial experiments, we employ the Neighbor Detail Template and Vicuna-7B. Additionally, we utilize Text-Graph Grounding Pre-trained GNNs \cite{tang2024graphgpt} (details are in Appendix \ref{sec:appendix_gdt}) as the graph description template and LLaMA3-8B \cite{touvron2023llama} as the base LLM model to evaluate our approach, separately. We compare the performance against the base model, which excludes the Stage 1 and category prompts in Stage 2, relying solely on task-specific tuning in Stage 2. Figure \ref{fig:compatibility} presents the results for Cora and Pubmed datasets. The findings indicate that our GALLM boosts model performance across different templates and base models, demonstrating the compatibility of our method as a universal approach.

\section{CONCLUSION}
In this paper, we propose GALLM, a novel LLM-based graph learning model that integrates Large Language Models into graph learning. We identify the shortcomings of existing methods, particularly their reliance on supervised tuning and ineffective self-supervised tuning. To address this, we align task templates for both stages. In the self-supervised tuning stage, we introduce a text matching task to align with the downstream tasks. In the task-specific tuning stage, we use category prompts to enhance alignment and help LLMs understand the task. This approach allows the model to benefit from self-supervised tuning, improving performance in downstream tasks. Our experiments demonstrate GALLM's potential as a graph foundation model, with exceptional performance across various scenarios. Future work will focus on incorporating more graph tasks and datasets to enhance the model's generalizability.

\clearpage
\bibliographystyle{ACM-Reference-Format}
\bibliography{main}










\section*{Appendix}
\appendix
\section{Experimental Setup Details}
\subsection{Baseline Descriptions}
\label{sec:appendix_baseline}
To verify the effectiveness of our proposed model, we compare it with various models, which can be divided into four groups:  traditional GNN models, transformer-based graph models, self-supervised GNN models, and LLMs for graphs. The first group includes:
\begin{itemize}[leftmargin=*]
    \item \textbf{GAT} \cite{velivckovic2017graph}: This method proposes an attention mechanism to aggregate neighborhood information.
    \item \textbf{GCN} \cite{kipf2016semi}: This method achieves neighborhood information aggregation by spectral graph convolutions.
    \item \textbf{GraphSage} \cite{hamilton2017inductive}: This method learns nodes embedding inductively based on the nodes’ ego features and the fixed number of neighborhood features.
    \item \textbf{SGC} \cite{wu2019simplifying}: This method simplifies the graph convolution operation by removing nonlinearities and collapsing weight matrices, all the while maintaining strong performance.
\end{itemize}
The second group includes the graph models based on the transformer network:
\begin{itemize}[leftmargin=*]
    \item \textbf{UniMP} \cite{shi2020masked}: This method adopts a Graph Transformer network to incorporate feature and label propagation at both training and inference time.
    \item \textbf{NodeFormer} \cite{wu2022nodeformer}: Against graph inconsistency, this method filters dissimilar neighbors based on a fixed threshold while aggregating multi-view neighborhood information.
\end{itemize}
The third group includes the self-supervised GNNs:
\begin{itemize}[leftmargin=*]
    \item \textbf{GraphCL} \cite{you2020graph}: This method introduces various graph augmentation techniques to derive diverse graph views, subsequently learning representations by maximizing feature consistency across these distinct augmented views.
    \item \textbf{SimGRACE} \cite{xia2022simgrace}: Without graph augmentation techniques, this method perturbs GNN encoders to obtain different views for contrastive learning.
\end{itemize}
The fourth group includes stage-of-the-art Large Language Models for graphs:
\begin{itemize}[leftmargin=*]
    \item \textbf{GraphGPT} \cite{tang2024graphgpt}: This method encodes graph data with a pre-trained GNN and proposes dual-stage instruction tuning to align graph data with natural language. 
    \item \textbf{LLaGA} \cite{chen2024llaga}: This method proposes two graph description templates to translate graph data and instruction tuning LLMs with three different tasks.
\end{itemize}
For zero-shot ability investigation, we further include a model focus on zero-shot performance:
\begin{itemize}[leftmargin=*]
    \item \textbf{ZeroG} \cite{li2024zerog}: This method fine-tunes language models to encode node attributes and class semantics, enabling cross-dataset zero-shot transferability.
\end{itemize}
For these baseline models, we implement them with the source code provided by authors or other researchers \footnote{https://github.com/pyg-team/pytorch\_geometric}. 

\subsection{Implementation Details}
\label{sec:imple_datail_appendix}
We employ Vicuna-7B-v1.5-16K as our foundation base model and utilize a pre-trained Sentence-BERT \cite{reimers2019sentence} to encode the nodes' raw text features. We set the batch size to 2 per device and the learning rate is $2e^{-3}$. The warm ratio is set to $3e^{-2}$ and the maximum input length of LLMs is 2048. We tune models for 3 epochs in both two stages. For the Neighbor Detail Template, we set the fixed sample size to 10 in each hop. In the text matching task, for each positive text sample, we sample 10 negative samples, consisting of 3 hard negative samples and 7 easy negative samples. In the category-enhanced soft prompt tuning, we utilize 3 virtual soft prompt tokens for each category. All the experiments are conducted with CUDA 12.1 and torch 2.3.1 on the NVIDIA H800 GPU.

\begin{table*}[t]
\caption{Instruction prompt of different datasets.}
\begin{tabular}{c|p{13cm}}
\toprule
 Dataset domain  & Instruction prompt  \\ \hline 
 Citation dataset & Given a node-centered graph: <node sequence>, each node represents a user. The first token represents the central node of the subgraph. The remaining represent the neighbors. we need to classify the center node into \{num\_labels\} classes: .... Please tell me which class the center node belongs?     \\ \hline
 Social network dataset & Given a node-centered graph: <node sequence>, each node represents a paper. The first token represents the central node of the subgraph. The remaining represent the neighbors. we need to classify the center node into \{num\_labels\} classes: .... Please tell me which class the center node belongs? \\
\bottomrule
\end{tabular}
\label{tab:prompt_input}
\end{table*}

\begin{table*}[t]
\caption{Explanatory texts for category manual prompts on node classification.}
\begin{tabular}{c|c|c}
\toprule
 Dataset & Category   & Explanatory text   \\ \hline 
\multirow{7}{*}{Cora} 
& case based  &  which analyzes real-world cases to draw conclusions     \\ \cline{2-3}
& genetic algorithms  & which applies evolutionary algorithms for optimization    \\ \cline{2-3}
& neural networks  &  which focuses on artificial neural networks and deep learning    \\ \cline{2-3}
& probabilistic methods &  which uses probability theory for modeling and prediction    \\ \cline{2-3}
& reinforcement learning &  which learns through trial-and-error interactions with the environment   \\ \cline{2-3}
& rule learning &  which extracts rules or logical expressions from data    \\ \cline{2-3}
& theory & which provides mathematical foundations and theoretical analysis \\ \hline
\multirow{3}{*}{Pubmed}
& diabetes mellitus type1 & which focuses on autoimmune destruction of insulin-producing cells \\ \cline{2-3}
& diabetes mellitus type2 & which focuses on insulin resistance and impaired glucose metabolism \\ \cline{2-3}
& diabetes mellitus experimental & which focuses on experimental studies for understanding and managing diabetes \\ \hline
\multirow{2}{*}{Instagram}
& normal & which engages with friends and shares personal moments \\ \cline{2-3}
& commercial  & which promotes products and services for business purposes \\
\bottomrule
\end{tabular}
\label{tab:prompt_mp}
\end{table*}

\begin{table*}[t]
\caption{Explanatory texts for category manual prompts on link prediction.}
\begin{tabular}{c|c}
\toprule
Category   & Explanatory text   \\ \hline 
connected  &  which means a paper citation relationship exists    \\ \hline
unconnected & which means they don't have a paper citation relationship  \\ 
\bottomrule
\end{tabular}
\label{tab:nc_et}
\end{table*}
\label{tab:prompt_lp}


\subsection{Dataset Details}
\label{sec:dataset_appendix}
We choose four representative widely-used datasets: Cora \cite{yang2016revisiting}, Pubmed \cite{he2023explanations}, Arxiv \cite{hu2020open} and Instagram \cite{kim2020multimodal}. These datasets originate from different domains, with the first three derived from citation networks and the last from social networks. 

For citation network datasets, nodes represent academic papers and edges denote citation relationships. Each node is attributed with a title and an abstract and is categorized into distinct research topics. 
Specifically: (1) the ArXiv dataset includes papers from the computer science arXiv website, categorized into 40 subject areas; (2) the PubMed dataset comprises papers from the PubMed database, categorized into 3 topics: experimentally induced diabetes, type 1 diabetes, and type 2 diabetes; (3) the Cora dataset includes papers related to artificial intelligence, labeled with 7 classes. 

For Instagram, nodes represent users, and edges signify ``following'' relationships. Users are attributed with follow lists, personal introductions, and tags, and are categorized into two classes: normal users and commercial users. 

\section{Details of Other Graph Description Template}
\label{sec:appendix_gdt}
In the Compatibility Investigation in Section \ref{sec:compat}, we replace the graph description template with Text-Graph Grounding Pre-trained GNNs proposed by GraphGPT \cite{tang2024graphgpt}. Firstly, this template samples a subgraph for the given central node to obtain the node sequences, where the central node is the first token and the neighboring nodes make up the remaining tokens. Then it employs a pre-trained GNN model to encode the graph tokens. Specifically, this template adopts a text-graph grounding paradigm, aligning graph features with their corresponding textual representations through contrastive learning. In this way, the graph information can be converted into pre-aligned token sequences.

\section{Complete Instruction Prompts and Explanatory text}
\label{sec:full_prompt_appendix}
In this section, we present our instruction prompts. Table \ref{tab:prompt_input} displays the prompts utilized for node classification on citation datasets and the social network dataset Instagram. Table \ref{tab:prompt_mp} provides the explanatory texts for category manual prompts used in task-specific tuning for node classification tasks. Due to the numerous categories in the Arxiv dataset, here we only show texts for the Cora, PubMed, and Instagram datasets. Additionally, Table \ref{tab:prompt_lp} outlines the explanatory texts for citation datasets in the link prediction task.

\end{document}